\documentclass[default,iicol,sn-mathphys]{sn-jnl}% Default with double column layout

\jyear{2023}%

\theoremstyle{thmstyleone}%
\theoremstyle{thmstyletwo}%

\theoremstyle{thmstylethree}%

\usepackage{diagbox}
\usepackage{ulem}
\usepackage{soul}

\begin{document}

%\title[Article Title]{Put fancy title here}
\title[Article Title]{The human intention. A taxonomy attempt and its applications to robotics}


\author*[1,2]{\fnm{J. E.} \sur{Dom\'{i}nguez-Vidal}}\email{jdominguez@iri.upc.edu}

\author[1,2]{\fnm{Alberto} \sur{Sanfeliu}}\email{alberto.sanfeliu@upc.edu}

\affil*[1]{\orgname{Institut de Robòtica i Informàtica Industrial},
\orgaddress{\street{Llorens i Artigas 4-6}, \city{Barcelona}, \postcode{08028}, \country{Spain}}}

\affil[2]{\orgname{Universitat Politècnica de Catalunya}, 
\orgaddress{\street{Jordi Girona, 31}, \city{Barcelona}, \postcode{08034}, \country{Spain}}}


\abstract{Despite a surge in robotics research dedicated to inferring and understanding human intent, a universally accepted definition remains elusive since existing works often equate human intention with specific task-related goals. This article seeks to address this gap by examining the multifaceted nature of intention. Drawing on insights from psychology, it attempts to consolidate a definition of intention into a comprehensible framework for a broader audience. The article classifies different types of intention based on psychological and communication studies, offering guidance to researchers shifting from pure technical enhancements to a more human-centric perspective in robotics. It then demonstrates how various robotics studies can be aligned with these intention categories. Finally, through in-depth analyses of collaborative search and object transport use cases, the article underscores the significance of considering the diverse facets of human intention.}

\keywords{Human-Robot Interaction, Intention Understanding, Human-centered Studies, Theory of Mind}



\maketitle

\section{Introduction}\label{sec:introduction}

In the dawn of robotics, the term robot was applied to automata oriented to perform simple and repetitive tasks to relieve the human from performing them. These early robots sensed the environment and acted accordingly~\cite{albus1975, brooks1986}. As they proved their usefulness and effectiveness, they went on to perform increasingly complex tasks~\cite{kam1997}, including increasingly frequent interactions with humans~\cite{terveen1995, triesch1998}, until today when they perform fully collaborative tasks requiring sophisticated systems that integrate a variety of sensors, actuators, and algorithms.

Some authors~\cite{sakita2004, duchaine2009, wakita2011, vernon2016, lyons2014} indicated the importance of understanding human intent in order to make these interactions safe for the human, reliable, comfortable, easily understandable, and the outcome of the interactions as productive as possible. Because of this, the last decade has seen an explosion of works that seek to infer, understand and even predict human intent.

However, if one analyzes these works, none of them provides a clear and general definition of intention, not even those works oriented to perform an overview and classification of related works~\cite{thomaz2016, losey2018}, but rather the human's intention is considered in terms of the task that the human is performing with the robot~\cite{maroger2021, thobbi2011, alevizos2020, alyacoub2021, nemlekar2019, wang2021, huang2016}. Thus, the human's intention is the trajectory~\cite{alevizos2020, maroger2021, thobbi2011} or velocity profile~\cite{alyacoub2021} they wish to follow, the place where they wish to deliver an object~\cite{nemlekar2019} or the next object they will select~\cite{wang2021, huang2016} among other possibilities. Because of this, it is possibly worthwhile to pause for a moment and ask ourselves the following questions. What is intention? Is it the same as a desire? Are intention and intentionality equivalent? Is there a single type of intention or can it be divided into multiple categories?

In this article we attempt to answer some of these questions. Based on the insights offered by psychology in recent decades, we attempt to combine and bring together various definitions of intention in a way that is easy to understand for the lay reader. Having done this, we present different types of intention that can be found in various psychological and interpersonal communication studies using diverse criteria including the degree of consciousness, temporality or type of goal, with the aim not to be exhaustive but to serve as a compass for other researchers who wish to move away from the purely technical enhancement of their work and towards the perspective of the robot's human companion. Subsequently, we show how multiple works present in the field of robotics can fit into the different categories presented previously. Finally, we analyze in detail two use cases so that the usefulness of taking into account and analyzing the polyhedral nature of human intention can be observed. Having done this, we also discuss other perspectives rarely explored in the literature and that we believe will pose some of the challenges for collaborative robotics in the years ahead.

We believe that this work can broaden the way we interpret a human-robot interaction by encouraging questions that are often overlooked and whose answer can open new perspectives in the development of collaborative robotics systems that are more responsive to human needs and desires.

In the remaining of the article, we start showing the definition of intention present in the psychology literature in Section~\ref{sec:definition}. In Section~\ref{sec:classification} we present a preliminary taxonomy with the most common categories of human intention. In Section~\ref{sec:applications} we present how these categories fit with multiple works in robotics. In Section~\ref{sec:use_cases} we analyze the use cases of collaborative search and collaborative object transport and in Section~\ref{sec:challenges} we link these examples with some future challenges. Finally, in Sections~\ref{sec:discussion} and~\ref{sec:conclusions} we present a brief discussion and the conclusions of the article respectively.

\section{Definition of human intention}\label{sec:definition}

The term intention includes multiple mental processes related to the information processing sequence carried out by our brain converting desires and goals into concrete behaviors. This preparation of the brain for action has been associated with the concept of 'free will' since the origins of philosophy. Descartes, for example, proposed that the mind, through its pineal gland, is able to consciously and freely choose between different actions and subsequently cause the body to execute the selected action. However, this dualistic perspective (mind-body separation) goes against the experimental results obtained by psychology and neuroscience in the last decades, causing conscious experience to be considered as a consequence of brain activity and not a cause.

\begin{figure}[t]
    \centering
    \includegraphics[width=0.93\linewidth]{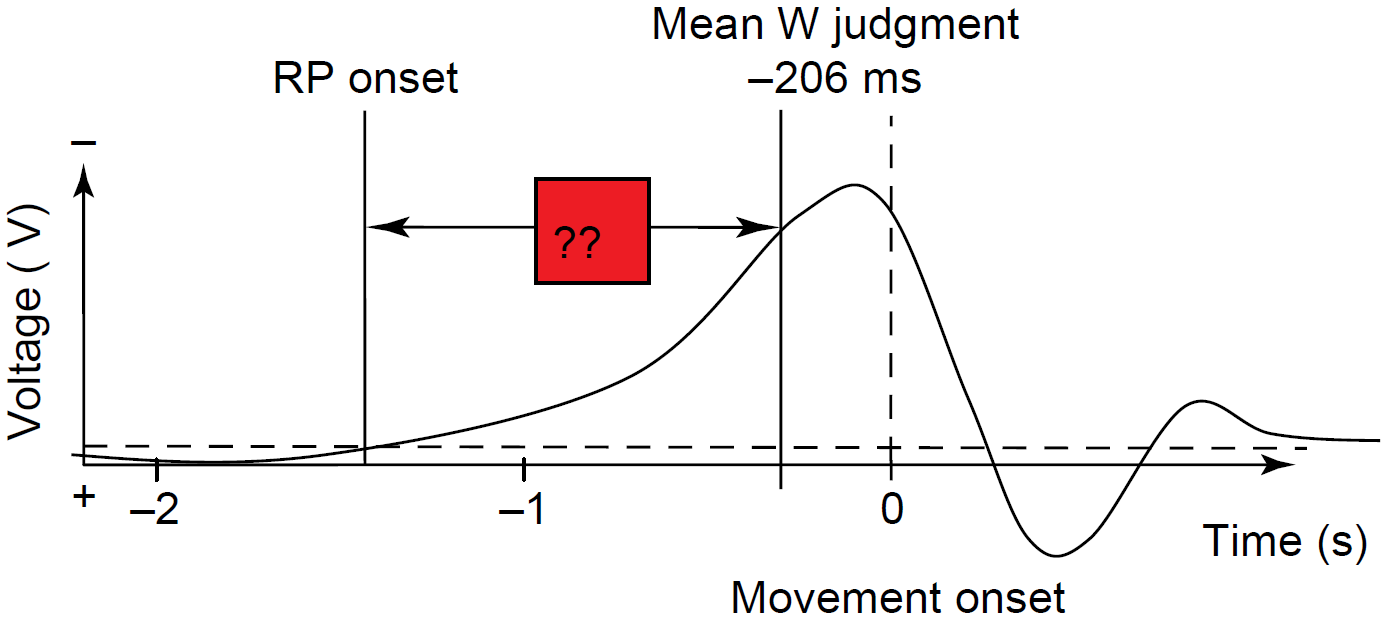}
    \caption{{\bf Representation of the evolution of the brain potential measured before and after the volunteer performed an action.} W marks the moment indicated by the volunteer as the one at which they decide to act. RP marks the moment at which the detected potential exceeds the average threshold that is considered an activation. Figure from~\cite{haggard2005}.}
    \label{fig:Libet_experiment}
\end{figure}

In this regard, experiments by Libet et al.~\cite{libet1983, libet1985} and later studies~\cite{lau2004, haggard2005} indicate that the initiation of an action involves an unconscious neural process, which subsequently produces the experience of intention. Specifically, Libet's experiment consisted of asking a series of volunteers to observe a moving dot on a screen and, when they wished to stop it, to perform a specific action (a slight movement of their right hand). Subsequently, the volunteers were asked to indicate where the dot was when they felt the need to act. The mean time between that moment and when the dot stopped is what Libet called "W judgement" (relative to 'free will'). However, Libet also measured the neural activity before and after the action performed by the volunteers, detecting that the brain began to prepare to act several hundred milliseconds before the moment when the volunteers considered that they decided to act. He termed this time "Readiness Potential" (see Fig.~\ref{fig:Libet_experiment}). Intention can therefore be regarded as the result of brain activity, rather than the cause of that activity.

\begin{figure}[t]
    \centering
    \includegraphics[width=0.9\linewidth]{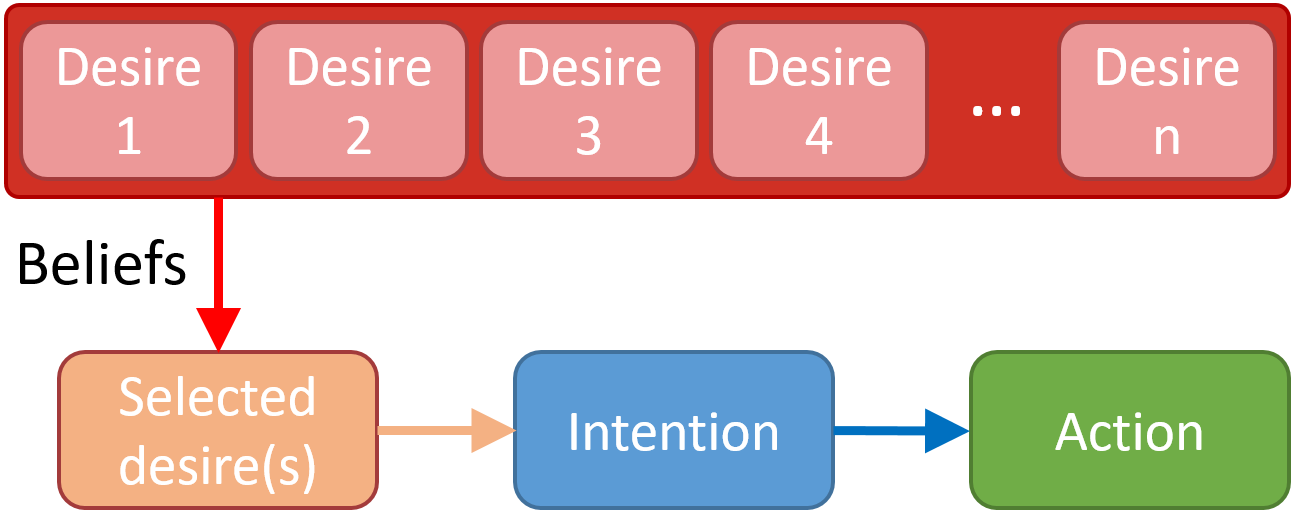}
    \caption{{\bf Relationship between desires, beliefs and intention.} The person's beliefs (based on their knowledge, experience, memory, etc.) cause them to end up choosing a particular desire which is the one that can materialize an intention.}
    \label{fig:Desire_Belief_Intention}
\end{figure}

Considering the above, some authors define intention as the "representation of the will to act"~\cite{fishbein1977}, the "determination to act"~\cite{shultz1980}, a "choice with commitment"~\cite{cohen1990}, or the "desire to achieve a result believing that a certain action can generate that result"~\cite{malle1997}. This is because a human can have multiple desires at the same time, even contrary to each other. However, at some point in time they must choose a subset of them, potentially just one, discard the others, and commit to planning and performing the actions necessary to satisfy them (see Fig.~\ref{fig:Desire_Belief_Intention}). Thus, intention emerges as a way to allow the agent to (1) observe their desire as a problem to be solved, (2) possess an 'admissibility screen' to rule out other counterproductive intentions, and (3) track the progress of their actions~\cite{bratman1987}. For example, if a human wishes to eat something, opting to cook an omelet makes them rule out going to a restaurant. The intention to make an omelet, in turn, allows them to (1) plan the ingredients they need, (2) discard the possibility of accompanying it with some fried eggs if they do not have enough eggs available, and (3) observe their actions and their progress toward making an edible omelet.

Thus, the concept of intention must be differentiated from that of desire. For some authors~\cite{forguson1989, malle1997}, the emergence of an intention requires that the human desires an outcome and has the belief that certain behaviors will lead to certain outcomes. That is, desire is prior to intention but by itself is not sufficient to generate the intention that produces an action~\cite{bratman1987what_is_intention}. The simplest example is "I wish I would win the lottery, but I have no intention of playing it". Thus, the intention to achieve a certain goal is usually the end result of deliberation over multiple wishes and desires in the pre-decision phase~\cite{Gollwitzer1996}. Thus, desires are often less connected to concrete actions and are more long-term oriented than intentions. Likewise, desires tend to include less feasible actions than actions associated with intentions because they have not yet been sifted by the agent to keep those that are really feasible~\cite{perugini2004}. 

At the same time, the relationship between intention and belief has been explored in multiple models that seek to explain human behavior. Among them, we can highlight the work of Fishbein and Ajzen~\cite{fishbein1977, ajzen2012, ajzen1985, ajzen1991}. According to the Theory of Reasoned Action~\cite{fishbein1977, ajzen2012}, a person's behavior is determined by their intention to perform a specific action. This intention, in turn, is based on two main factors: the attitude toward the action and the subjective norm. This attitude towards the action can be positive or negative and is influenced by the beliefs that that person has about the consequences of the action and the valuation they give to those consequences. The subjective norm, on the other hand, refers to the belief that a person has about the expectations of others regarding their behavior. This theory is extended in~\cite{ajzen1985, ajzen1991} by including the concept of 'perceived control', which refers to a person's belief about how much control they have (availability of resources, constraints, etc.) over the performance of the behavior in question. Thus, the person's beliefs determine how strong their intention to perform a particular action will be.

It is worth mentioning that in~\cite{ajzen2005} the author himself acknowledges that "The theory of Reasoned Action was developed explicitly to deal with purely volitional behaviors"; i.e., simple behaviors, where successful performance of the behavior requires only the formation of an intention~\cite{armitage2000}. This is due to the existence of the so-called intention-behaviour gap~\cite{sniehotta2005, bhattacherjee2009, mohiyeddini2009} and explains that having the intention to perform an action does not necessarily lead to the performance of an action, or more specifically, the intended action.

\begin{figure}[t]
    \centering
    \includegraphics[width=0.95\linewidth]{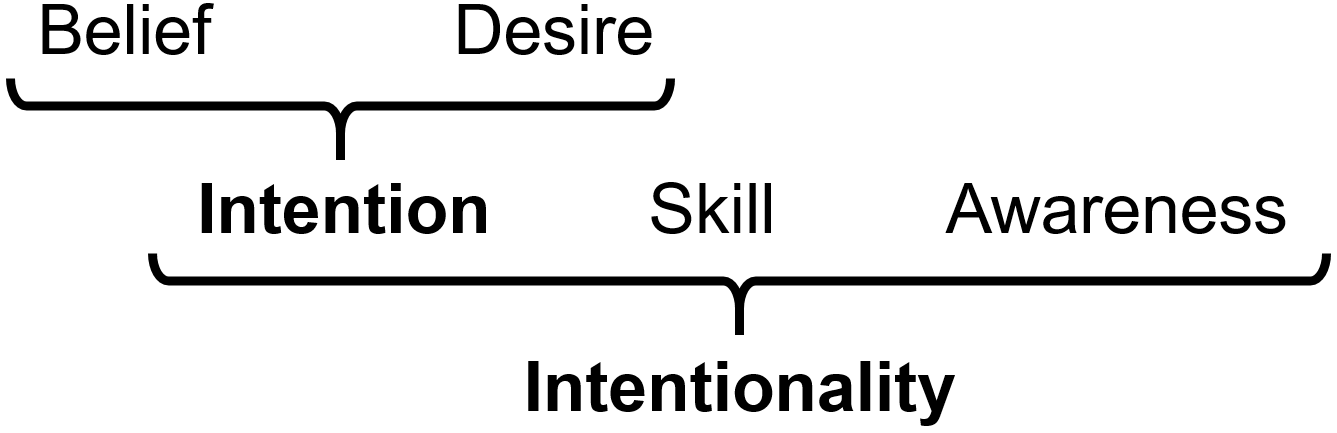}
    \caption{{\bf A model of the relationship between the intention and intentionality concepts.} To have an intention is a pre-requisite to have an intentionality. Figure based on concepts presented in~\cite{malle1997}.}
    \label{fig:intention_vs_intentionality}
\end{figure}

This difference between intention and intentionality is discussed in~\cite{malle1997}. Through various user studies with which the authors seek a definition of intentionality, they consider that there are two further requirements for acting intentionally. First, a minimum of conscious awareness of fulfilling the intention while performing the action. If you intend to call someone, then you remember that you should call someone else first, but by mistake you dial the number of the first person; no doubt you intended to call them, but you did not end up doing so intentionally. Secondly, they consider that it is also necessary to possess the necessary ability to perform the action intentionally. If it is the first time you play darts and in your first throw you get the maximum score by chance, it is clear that you intended to get it but you did not do it intentionally. Thus, intention is a necessary but not sufficient condition for intentionality (see Fig.~\ref{fig:intention_vs_intentionality}).

\section{Classification of human intention}\label{sec:classification}

If we accept the above definitions, they are still sufficiently loose to allow human intention to manifest itself in different ways. Below are some possible classifications that can be found in the literature.
It is worth mentioning that some of the distinctions to be discussed are treated with the same terms as those used in this article (goal-oriented VS. implementation intention or individual VS. shared intention) while other possible classifications arise by combining perspectives of different authors from different disciplines (including psychology or interpersonal communication) who even use different terms to refer to the same concepts. On the other hand, not all the works on which the following classifications are based take into account the difference between intention and intentionality discussed above. This error will be accepted for the sake of bringing together multiple works under the same comparative.

\subsection{Goal-oriented intention VS. implementation intention}

Having the intention to call someone is not the same as having the intention to get up, go to the phone and dial someone's number. The first case corresponds to the goal-oriented intention to realize a concrete goal while the second case would be an example of the intention to implement the previous goal-oriented intention~\cite{gollwitzer1993, gollwitzer1999, gollwitzer1997}. While goal-oriented intention responds to the form "I intend to do X", implementation-oriented intention follows the structure "I intend to do X in situation Y"~\cite{sheeran2002}.

The first intention that occurs is the goal-oriented intention, "I am going to call X". Once this intention appears, the human plans how to carry it out and then discovers that they need to perform an action prior to the one that would allow them to fulfill their intention, in this case, to approach the telephone because it is not in their proximity. Thus, the implementation intention arises, which includes actions that are only necessary due to the specific situation in which the human finds themselves, making their implementation intention to be "I am going to get up and go to the phone to call X". Using one of the examples from the previous section, the goal-oriented intention would be "I am going to make an omelet" while possible implementation intentions would be "I am going to peel potatoes before lighting the fire" or "I am going to whisk eggs while the oil is heating". Trying to translate this into robotics terms, we could say that the goal-oriented intention is equivalent to the goal of a global planner and the implementation intention is the sequence of sub-goals of a local planner.

Note that these two intentions are neither opposite nor exclusive. Although the goal-oriented intention is generally formed earlier, if both exist, they do so at the same time. In fact, the goal-oriented intention is the one that allows to initially choose the goal to be achieved in a first stage and the implementation intention is the one that allows to visualize how to accomplish that goal making it more feasible~\cite{gollwitzer1993, gollwitzer2006}. Additionally, the implementation intention is by nature more concrete in time and space which makes a goal-oriented intention accompanied by an implementation intention more likely to produce actual behavior~\cite{gollwitzer1997, verplanken1999, sheeran2000}. 

Related to the above, in the literature one can find the difference between intrinsic desires (oriented to achieve an end state) and extrinsic desires (oriented to execute a series of actions designed as a means to achieve an end state)~\cite{mele1995}. This may open the door to intrinsic and extrinsic intentions. However, the terms found and widely accepted are the aforementioned goal-oriented and implementation intentions.

\subsection{Implicit intention VS. explicit intention}\label{sec:implicit_VS_explicit}

Human intention is the product of a neural process in which one among multiple paths of action is chosen so that the human can focus on a single possibility, plan about it and assume its consequences. However, from the point of view of an external observer, this process is not observable. Moreover, an agent's actions may be interpreted differently by the agent performing them and by an outside observer~\cite{jones1987}. Therefore, in order for a human B to know the intention of a human A, either B must infer it from A's actions or A must explicitly communicate it to B~\cite{sartori2011, yus1999}.

Thus, from the point of view of an external observer, a distinction can be made between implicit or inferable intention and explicit intention~\cite{park2016}. The former alludes to those inferences that this observer can make about the intention of another human by observing their actions. If we see a person waiting at a crosswalk looking attentively at a traffic light, we can deduce that their intention is to cross to the other side. In turn, implicit intention also refers to those intentions related to an interaction between two humans in which one of those humans does not express them directly because they trust that the other is capable of inferring them, either because they consider them to be the obvious result of shared information, is embarrassed to express them, or simply because they are intentions associated with actions to be carried out in such a short period of time that the time cost associated with expressing them explicitly is prohibitive~\cite{sperber1986, yus1999}. It is worth mentioning that there is a further motivation for not expressing explicitly all the information that is considered relevant, and that is to give the other agent time to react and thus be able to adjust the mental model that one has of the other agent. This is more common between two humans who have just met or between a human and a robot when the human is learning the robot's capabilities.

On the other hand, explicit intent refers to the one the human explicitly expresses to another human, whether using natural language, gestures or any other type of code agreed upon by both and unequivocally understandable to both parties. Its use serves to minimize or eliminate misunderstandings and/or reduce uncertainties~\cite{clark1996, yus1999}.

Note that the explicitly expressed intention may be contrary to and even exclusive of the implicit intention being shown at the same time. This type of situation can occur either due to the human's unawareness that their actions are not consistent with the intention they really have~\cite{burgoon2006, knapp2013}, or in situations in which they are resorting to deception~\cite{depaulo2003, vrij2010}. Examples of the latter are rehearsed plays in team sports. The positions of the players of a team may be indicative of one type of play and the play indicated by the captain or, in general, by the player in command, may be completely opposite.

\subsection{Conscious intention VS. unconscious intention}

It has been indicated that human intention arises after a preparation process from which the human chooses a desired goal and acts accordingly. However, this decision process is not always made consciously~\cite{wegner1998, palfi2021}. Just as the degree of information processing prior to the formation of an intention varies from one human to another depending on their motivation and cognitive abilities, so does the degree of awareness with which one intention is decided upon over another~\cite{ajzen1999, ajzen2015}.

Notable examples are brushing teeth or showering. While it is possible that both activities are performed deliberately due to the discomfort caused by a food remnant between the teeth or the presence of sweat after intense physical activity, both actions are generally performed routinely and without paying much attention. In any case, if you ask the human who has just performed either of these two activities, they will be able to remember the moment when they made the decision to perform them, even if they were not thinking about it at the time.

Thus, from the point of view of the degree of consciousness, conscious intention and unconscious intention can be distinguished. The former refers to those intentions of which the human is aware at the very moment of their formation, thus being in control of their actions~\cite{wegner1998}, while the latter refers to those intentions of which the human is only aware when they make the effort to ask themselves the reason for performing the corresponding action. Even though, unconscious intentions still allow them to fulfill the desired goal by adapting to the situational changes that may occur~\cite{bargh1999, bargh2001}. This second type of intentions are usually associated with habits, automatisms, routines and even manias~\cite{sheeran2013, ajzen2015}.

An example of both types of intentions occurring together may be the intention to "go to eat at a restaurant". Choosing the restaurant can be the result of a conscious intention if it is an occasional activity or an unconscious intention if it is done as a routine. The same is true for choosing the menu. On the other hand, asking for the bill once the meal is finished is usually an action associated with an unconscious intention that is rarely thought about consciously.

\subsection{Individual intention VS. collective intention}

The actions performed by an agent can be individual or collective. That is, they can be executed by/on themselves without the need to interact with another agent or, on the contrary, they can depend on the interaction with one or several external agents. In the first case, the intention that would motivate it can be referred to as individual intention. While in the second case, it would be a collective intention~\cite{searle1990, dunin2002}, social intention~\cite{becchio2008} or shared intention~\cite{bratman1993, gilbert2009} that is governed not only by the capacities, knowledge, beliefs and desires of the agent themselves but also by those of other agents with whom they must interact.

This collective intention is bidirectional, that is, it refers both to the intention of the original agent with the other agents and to that of the other agents with the first agent. It can arise spontaneously or explicitly by prior agreement and it will exist as long as the commitment between both agents to work together for the achievement of shared goals is maintained~\cite{gilbert2006, gold2007}. It is this intention that turns the expression "I and you are going to do X" into "we are going to do X"~\cite{gilbert2009, tuomela1988} and it is this intention that allows to (1) coordinate the actions of both agents so that both pursue the common goal, (2) coordinate the plans of each agent so that they both conform to the roles that both agents play to achieve the common goal, and (3) provide a common framework that allows structuring negotiations since different agents may have different preferences but their common goal forces them to negotiate~\cite{bratman1987, sadler2006, tomasello2005}.

This distinction between individual intention and collective intention provides insight into how joint actions are carried out in the context of social relationships. Collective intention implies mutual agreement and shared awareness~\cite{gilbert2003, gilbert2009}, whereas personal intentions focus on individual goals. This distinction is essential for understanding how people cooperate and coordinate in social situations and how social relationships based on joint action are formed and maintained. Indeed, it is this group intention shared among several agents that allows and encourages the emergence of negotiation (to check to what extent their goals are common)~\cite{thompson1990, dedreu2010} and arbitration (deciding who does what for the achievement of the common goals)~\cite{losey2018} processes.

Individual intention includes multiple of the above examples (going to eat at a restaurant, brushing one's teeth, calling to person X). Collective intention, on the other hand, can include both cooperative actions (going to eat with X at a restaurant) and competitive actions (any play in an individual sport such as tennis) or a combination of both (any play in a team sport such as soccer). Due to the above, according to some authors, collective intention can in turn be classified between cooperative intention and competitive intention~\cite{becchio2008b, sartori2011, manera2011}. Although the latter category may seem contradictory, the fact is that both agents coordinate their actions in function of those of the opponent and both assign each other a role, in this case, that of opponent or adversary.

While in the literature this group intention can be found as collective, shared or social intention, in this article we chose to use the term collective intention. This is because it was the original term that motivated the appearance of the following ones. In turn, the term social intention implies the assumption of social rules that may vary from one culture to another, restricting the breadth of the concept that this article seeks to present.

\subsection{Short-term intention VS. long-term intention}

From a temporal point of view, intention can also be classified as short-term intention and long-term intention, the former referring to the one which implies actions that are close in time and the latter to that which requires actions that are more distant in time. Because intentions that are close in time are more easily planned, since they depend on fewer variables and therefore have greater certainty as to their feasibility, they tend to be more likely to materialize into concrete actions than if they are planned in the long term~\cite{tulloch2009}, making them better predictors of human behavior.

The main example studied in the literature is the intention to "revisit a tourist destination". While the action is the same, it has been found that the likelihood of this occurring within a few months is determined by the degree of satisfaction with the previous visit, while in the long term it is highly dependent on the tourist's novelty seeking~\cite{jang2007, assaker2013}. 

We recognize that this is the least addressed classification in psychology as well as in related fields. However, for the sake of completeness, we consider it necessary to include it. Apart from the fact that it allows us to explain some phenomena in the field of robotics as it will be shown in the next section.

\begin{table}[t]
\caption{Summary of the different ways of classifying human intention according to the literature}
\centering
\label{table:intention-classification}
\begin{tabular}{lc}
\hline
Criterion        & Type of intention                        \\ \hline
Type of goal     & Goal oriented -- Implementation          \\
Communication    & Implicit -- Explicit                     \\
Consciousness    & Conscious -- Unconscious                 \\
Number of agents & Individual  -- Collective                \\
Temporality      & Short-term -- Long-term                  \\ \hline
\end{tabular}
\end{table}

The Table~\ref{table:intention-classification} presents a summary of the possible classifications analyzed in this section. It should be noted that the categories presented are not necessarily mutually exclusive. Furthermore, there are certain relationships among them that allow the same intention to be analyzed in different ways. For example, an explicit intention is by nature conscious and tend to be shared with the agent to whom this intention is expressed, but it can be either goal-oriented or implementation-oriented. At the same time, an implicit intention, which must be inferred by the external observer, can be either of an individual type if it does not involve any interaction with this observer or of a collective type otherwise. At the same time, it is often, but not necessarily, of an unconscious type. 

\begin{figure}[t]
    \centering
    \includegraphics[width=0.95\linewidth]{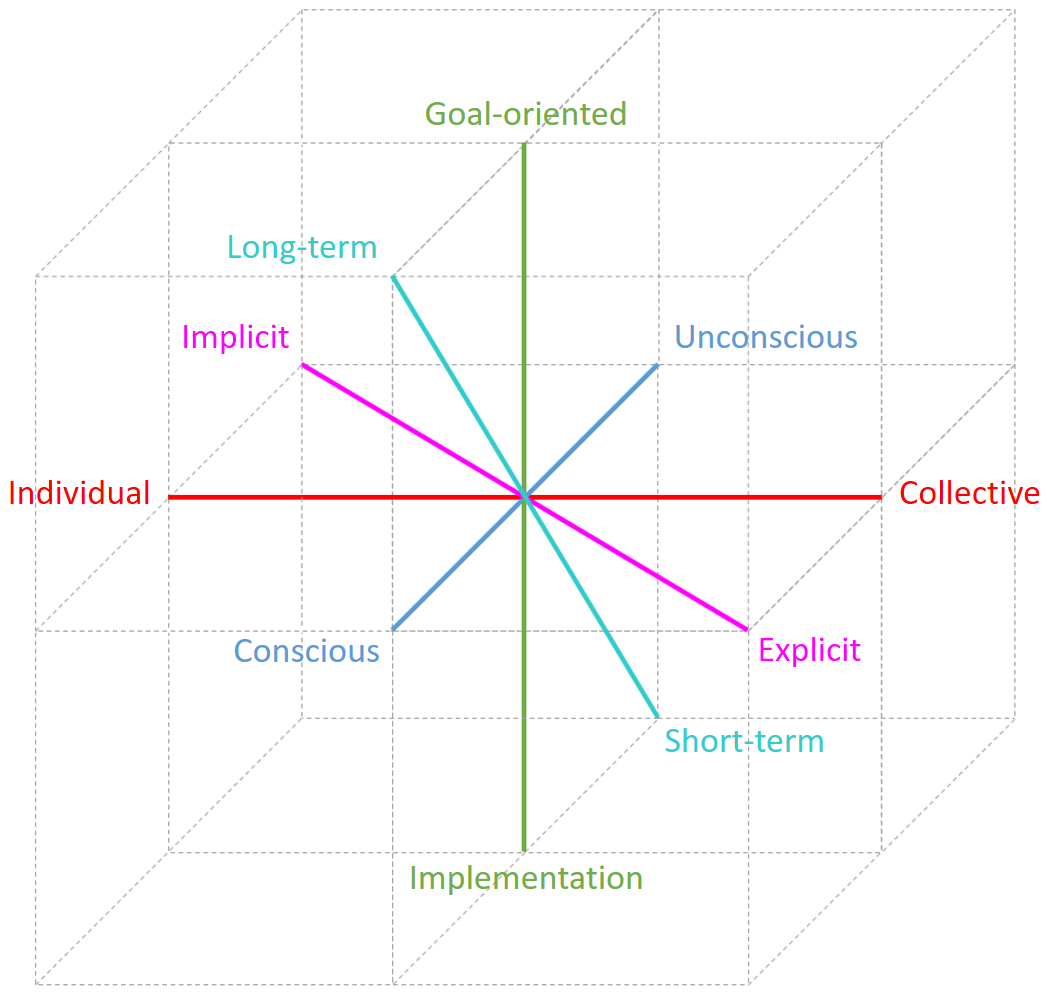}
    \caption{{\bf Representation of the main relationships between intention categories.}}
    \label{fig:categories-relation}
\end{figure}

Fig.~\ref{fig:categories-relation} tries to compress all the previous classifications in a single diagram showing some of the main relationships that can happen. For this purpose, a geometrical representation is used whereby those classifications that are, or tend to be, independent of each other are represented by perpendicular axes. Similarly, those classifications that are related are represented by axes which can be projected over the related classifications. For ease of visualization, the three primary colors are used for the independent classifications and colors obtained by mixing two primary colors are used for the dependent classifications. Thus, an explicit intention can be projected on the axis of shared intentions as well as on the axis of conscious intentions, but not on the axis of goal-oriented or implementation-oriented intentions, since an explicit intention can belong to either of these two categories, and the same is true of implicit intention. This same diagram also shows how a long-term intention tends to be of the conscious type and to be associated with goal-oriented intentions.

\section{Intention taxonomy applied to robotics}\label{sec:applications}

From the point of view of robotics, and to the best of our knowledge, intention has not been defined in the terms of Section~\ref{sec:definition}. Instead, it has been used the folk concept of intention, which is present, with slight differences, in the common imagination of all people~\cite{malle1997}. In this way, intention has been a concept used {\it ad-hoc} depending on the specific task being analyzed. In a hand-over task in which the human is the agent delivering the object, the human's intention is whether or not they wish to deliver the object~\cite{wang2018}, what the object delivery point is~\cite{nemlekar2019} or the trajectory to be described~\cite{zhang2020} depending on what is being studied in each particular research work. Similarly, in collaborative transport tasks, the human's intention is either the location to which the human wishes to transport the object~\cite{nicolis2018} or the desired trajectory~\cite{laplaza2022, mavridis2018, alevizos2020, ge2011, li2013} or the desired velocity profile~\cite{alyacoub2021}. In other works~\cite{park2019, liu2019, huang2016, piccarra2018}, the human's intention refers to the next object to pick up, the next ingredient to select, or whether or not the human wishes to collaborate with a robot. In some works, the human's intention is even categorized as a finite set of actions it can perform next~\cite{yu2015}. Moving away from human-robot collaboration, in autonomous robot navigation tasks in an environment with multiple humans present, each with their own goal, the human's intention has been considered as both the destination they wish to reach~\cite{ferrer2014} and the path they wish to follow~\cite{ferrer2014b}. Therefore, there is no unified or consensual criterion of what the human's intention is.

\begin{figure}[t]
    \centering
    \includegraphics[width=0.95\linewidth]{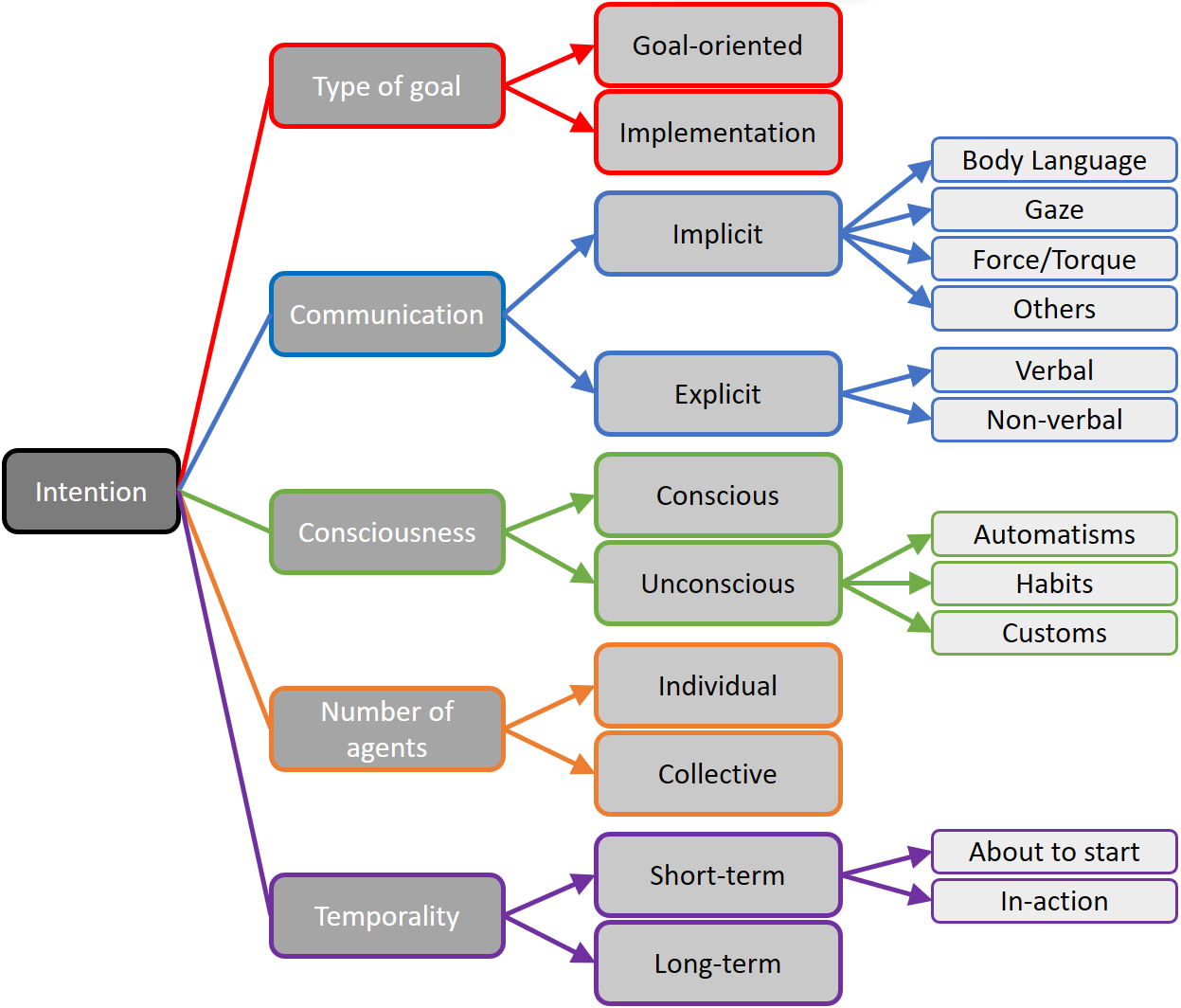}
    \caption{{\bf Outline of the taxonomy presented for human intent.}}
    \label{fig:taxonomy}
\end{figure}

Multiple of the above examples may simultaneously belong to several of the categories presented in the previous section even though no distinction was made in their respective articles. The following subsections will show how multiple works in the field of robotics can fit into one, or even several at the same time, of the categories discussed above. At the same time, Fig.~\ref{fig:taxonomy} shows a summary of the different types of intent considered in the present work based on the criteria in Table~\ref{table:intention-classification}. All these types of intention will be defined in the following subsections from the point of view of their current application in the field of robotics.

\begin{figure}[t]
    \centering
    \includegraphics[width=0.9\linewidth]{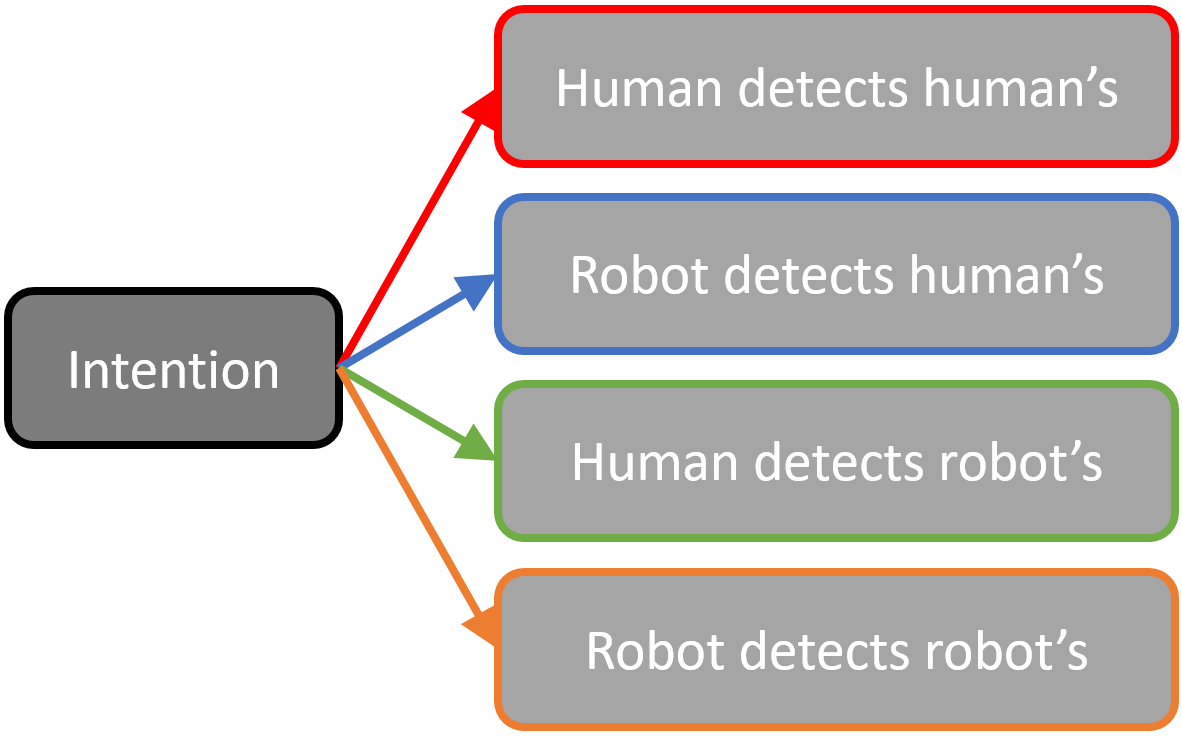}
    \caption{{\bf Different perspectives considered about who is the agent detecting the intention.}}
    \label{fig:taxonomy-point-of-view}
\end{figure}

It is worth mentioning that there is another point of view when analyzing the concept of intention, which is meaningless in psychology but does make sense in the field of robotics. This point of view consists of attending to who is the agent observing the intention of the other agent. Fig.~\ref{fig:taxonomy-point-of-view} shows the four possibilities considered. Psychology focuses on the study of the human mind. Because of this, all the studies presented in the previous sections would assume the first possible perspective, i.e., that it is a human the one that detects the intention of another human.

Robotics, on the other hand, has focused on providing the robot with capabilities to detect human intention. For this reason, the vast majority of works in the literature and, therefore, presented in this section, follow this second perspective, in which the robot must detect and interpret the human's intention in the best possible way. However, there are also the other two perspectives, much less explored but equally promising nonetheless. Because the number of articles in the literature that seek to analyze the latter two perspectives is notably lower, we prefer to leave their analysis for Section~\ref{sec:challenges} in which we present some of the future challenges that we think collaborative robotics may face in the future.

\subsection{Type of goal: Goal-oriented intention VS. Implementation intention}

Starting from the definition of intention offered in the physiologist work~\cite{malle1997}, i.e., "the desire to achieve a result believing that a certain action can generate that result"; we are going to use the following definitions in robotics:

\begin{itemize}
   \item \textbf{Goal-oriented intention:} the desire to achieve that result and not another believing that there is an action or sequence of actions that makes it feasible. 
   \item \textbf{Implementation intention:} the desire to perform a particular action or sequence of actions and not another believing that it makes it possible to achieve the desired result. 
\end{itemize}
   
Thus, we consider all of those works that define the human's intention as one among a finite set of goals or final states that they wish to achieve, would be understanding this human's intention as a goal-oriented intention. In turn, those works that define the human's intention as the human's way of reaching this goal or final state are making use of the definition of implementation intention. This second type of intention is usually more complex to understand or predict, because it can have quasi-infinite forms and values.

\begin{figure}[t]
    \centering
    \includegraphics[width=0.96\linewidth]{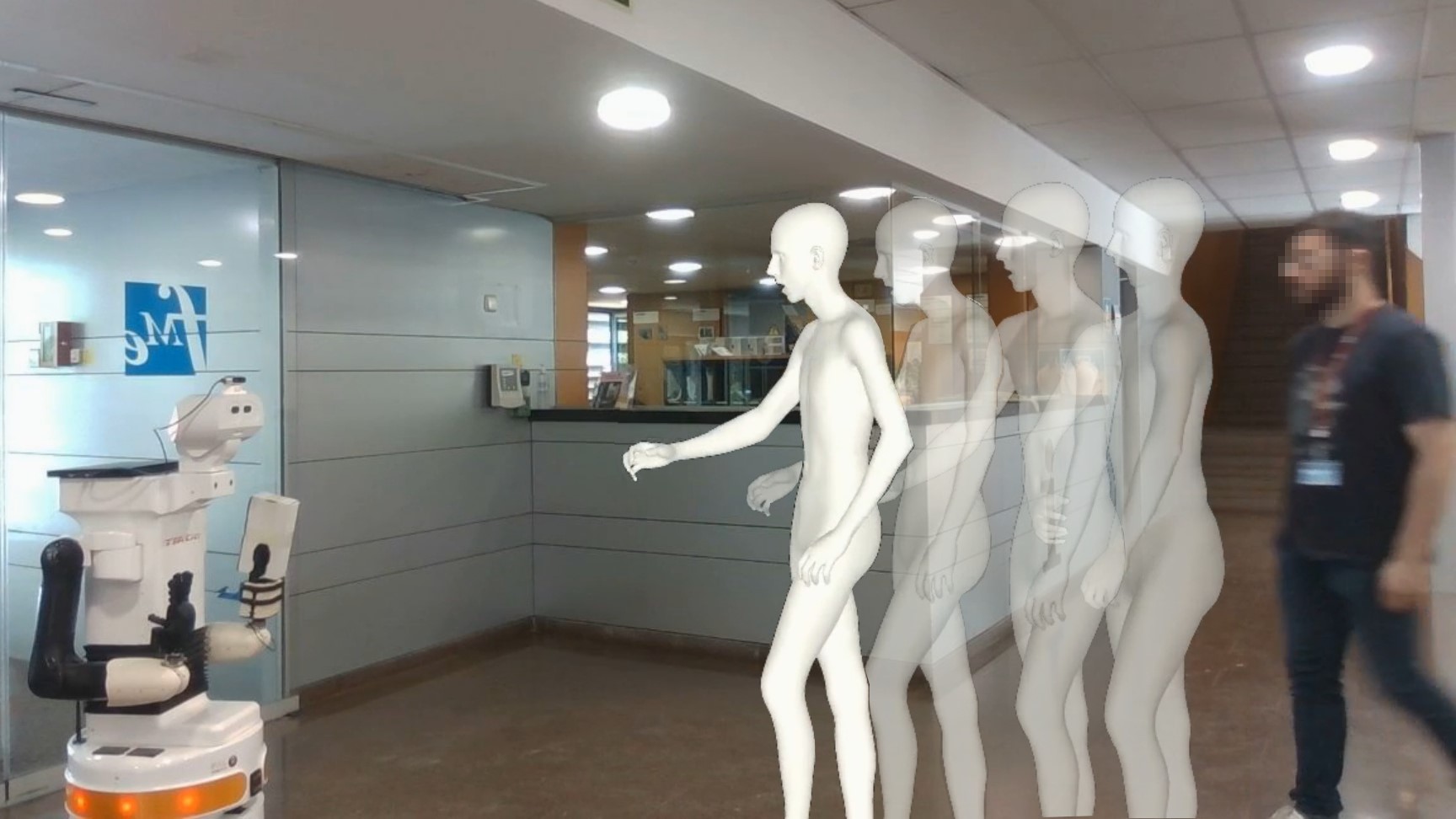}
    \caption{{\bf Handover example.} Using a trajectory predictor, the robot can predict how much the human is going to approach (implementation intention) and even the desired Object Transfer Point (goal-oriented intention) that the robot should reach.}
    \label{fig:handover}
\end{figure}

Regarding the specific task of hand-over,~\cite{wang2018} use a wearable based on electromyography (EMG) and inertial measurement units (IMU) sensors that they place on the human's arm to detect when the human wants or not to receive an object delivered by the robot based on the movement that the human performs. This would be a case of goal-oriented intention. So would be in~\cite{nemlekar2019}. In this case, they infer the Object Transfer Point (OTP), or place where the human wishes to receive or deliver an object. To do so, they precompute a static OTP based on giver's position, reachability and height and combine it with a dynamic OTP based on an estimation of the human's arm movement using Probabilistic Movement Primitives (Pro-MP). This second component, although used to estimate a final goal, if used individually, could be understood as an implementation intention. This idea of estimating "the trajectory that the human intends to follow" is the one used in~\cite{laplaza2022} using for that a multi-head attention Graph Convolutional Network (GCN) model which receives as input the images of a camera in the robot's head, understanding the human's intention in this case as an implementation intention (see Fig.~\ref{fig:handover}).

It is also possible to find works on collaborative object manipulation tasks in which the human's intention is understood as implementation intention, either referring to the trajectory that the human desires for the manipulated object, either using control-based~\cite{mavridis2018, alevizos2020} or Deep Learning models~\cite{ge2011, li2013}, or referring to the velocity profile that the human desires for the object~\cite{alyacoub2021}, in this case, using Learning from Demonstration (LfD) methodologies with Weighted Random Forest (WRF) as their workhorse. Similarly, the robot may try to infer the particular object that the human wishes to manipulate~\cite{jain2019}, this being a case of goal-oriented intention. If, in addition to manipulating, the human wishes to collaboratively transport the object, it is possible to find both the approach of predicting the location to which the human wishes to take the object~\cite{nicolis2018} (goal-oriented intention) and that of predicting the trajectory they wish to follow~\cite{Inferencia_ROMAN2023, Predictor_IROS2023} (implementation intention).

Finally, in autonomous navigation tasks, understanding the intention of each human the robot encounters as the "destination they wish to reach"~\cite{ferrer2014} would be a case of goal-oriented intention. Meanwhile, understanding their intention as "the path they wish to follow"~\cite{ferrer2014b} would correspond to an implementation intention.

In general, this distinction between goal-oriented and implementation intention allows us to focus on trying to understand first the human's goal-oriented intention and then proceed to understand their implementation intention, which is typically dependent on the former.

\subsection{Communication: Implicit intention VS. Explicit intention}\label{sec:implicit-explicit}

According to the definitions in the previous section, we are going to use the following definitions in robotics:

\begin{itemize}
   \item The \textbf{implicit intention} of agent A is the intention that must be inferred by agent B, the observer, by means of a reasoning mechanism and applied to the observation of the actions of agent A. 
   \item The \textbf{explicit intention} of agent A is the intention indicated by agent A to agent B by means of a communication code known and accepted by both. 
\end{itemize}

In this way, those works that attempt to predict or infer human intention from their actions, movements or gaze are alluding to the implicit type of intention. In this group we can include most of the predictors already discussed~\cite{ferrer2014b, nicolis2018, ge2011, li2013, alyacoub2021, park2019, liu2019, huang2016, piccarra2018, laplaza2022, Inferencia_ROMAN2023, Predictor_IROS2023, jain2019} among others. All of them make use of some kind of inference engine: Bayesian predictors, Support Vector Machines (SVM), Gauxian Mixure Models (GMM), Recurrent Neural Networks (RNN), GCN, or combinations of Convolutional Neural Networks (CNN) and Long Short-Term Memory (LSTM) units. At the same time, all of them usually use as input signals the previous human movement, either as a whole, distinguishing its most representative joints or focusing on the limb related to the task being performed. Another commonly used input signal is the human's gaze, which is really useful to determine where the human is focusing its interest~\cite{huang2016}. Other works~\cite{Mortl2012} use direct signals, for example force or torque, to indicate in which direction a table has to be moved.

Less frequent are the works that consider the explicit intention of the human, that is, the one directly given by the human, typically to avoid misunderstandings or because it is not possible to infer it otherwise. It is possible to find works that use natural language processing (NLP) to know the human's intention. For example, in~\cite{li2021, mi2020} they use a combination of visual and natural language processing so that the robot first recognizes the environment and then knows which is the object that the human really wants or which one best fits the request that the human is making. In turn, works based on human gesture recognition also make use of this definition of explicit intention. For example, in~\cite{peral2022, cucurull2023} they train different Deep Learning models to detect the gestures made by the human and, with them, to know what is the next action that the human expects from the robot. Both NLP and gesture recognition can be found in industrial environments when the robot is required to recognize quickly and minimizing possible misunderstandings about the real human's intention~\cite{mukherjee2022}. Similarly,~\cite{lorentz2023} combines both types of information to obtain the human's explicit intention about where they want the robot to move the desired object.

\begin{figure}[t]
    \centering
    \includegraphics[width=0.96\linewidth]{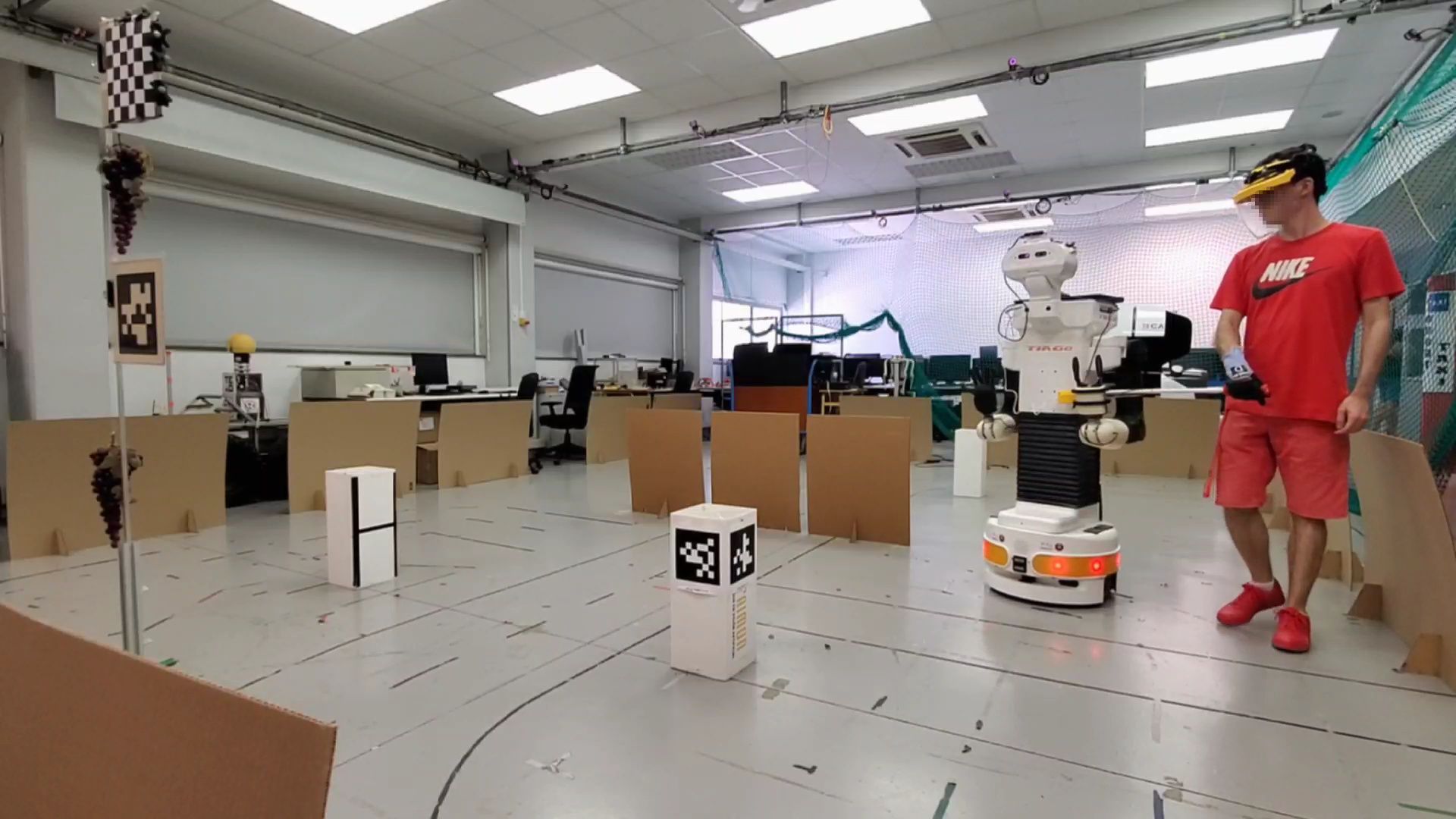}
    \caption{{\bf Human-robot collaborative object transportation example.} Human and robot transporting a bar. Force sensor attached to the robot's wrist with which it can  the human's implicit intention and buttons on the handle of the object for explicit intention communication.}
    \label{fig:collaborative_transportation}
\end{figure}

There are also works that take into account both types of intentions. In~\cite{huang2016}, the human's gaze is analyzed to predict the next object to be selected. Once the robot is sufficiently certain about it, it makes a plan to be able to grab that object but it does not execute this plan until the human verbally confirms their choice. Similarly, applied to collaborative transportation, in~\cite{PIA_HRI} the direction the human wants to go is estimated from the force they exert on the object, but the human is also allowed to communicate with the robot explicitly to avoid possible misunderstandings in complex situations (see Fig.~\ref{fig:collaborative_transportation}). Finally, there is also the opposite approach, i.e., that it is the robot that indicates its intention about the path it is going to follow either explicitly~\cite{chadalavada2015} or both implicitly and explicitly at the same time~\cite{che2020}.

In general, taking into account both types of intent and not focusing on only one of them helps to reduce uncertainties and achieve interactions that are more humane. This approach is tested in~\cite{Inferencia_ROMAN2023}, which shows that the human does not want the robot to infer everything, but also wants to be able to communicate with it.

\subsection{Consciousness: Conscious intention VS. Unconscious intention}

Based on the definitions in the previous section, we are going to use the following definitions in robotics:

\begin{itemize}
   \item The \textbf{conscious intention} is an intention whose formation occurs when the human is aware of it.
   \item The \textbf{unconscious intention} is an intention that happens without the human being aware of it.
\end{itemize}

Therefore, in this work we understand the use of unconscious intention as automatisms, habits and customs. As such, it has been used indirectly in two ways. On the one hand, by taking advantage of the human's unconscious gaze patterns to obtain clues as to what the human's real intention is. On the other hand, by learning their habits and customs to learn the human's preferences and thus make the robot act more comfortably.

In the first group, one can find works that explore how gaze helps in human-human tasks to improve their performance by allowing better understanding of ambiguous situations and then use this knowledge to modify the robot's gaze to help the human to understand what the robot wants to do~\cite{boucher2012, mutlu2009}. Similarly, in~\cite{aronson2018} they use glasses as an external gadget to analyze the human's gaze and with this detect what is the object of interest in a human-robot shared manipulation task. Using the same sensor, in~\cite{chadalavada2020} it is proven that by analyzing the human's gaze pattern it is possible to predict which way they will tend in order to dodge the robot if they encounter it in a narrow corridor. They also found that the signal used by the robot to indicate its trajectory affects the gaze pattern and that this phenomenon can be exploited to make it easier to detect the human's intention. Also applied to autonomous navigation,~\cite{zhang2023} uses the gaze of the people present to know if they are paying attention to the robot and therefore the robot can move faster knowing that they are aware of its presence.

In the second group, works like~\cite{sisbot2012} modify the robot's planner so that the robot adapts to the human's vision field, posture, and preferences so that the human receives the object delivered by the robot in the most comfortable way. Similarly,~\cite{yang2020} performs a classification of possible human hand grasp types in hand-over tasks so that the robot takes into account the human's preference when delivering or receiving an object.

Examples of works that understand human intention as conscious intention would be all of those previously discussed in which the human explicitly expresses their intention~\cite{li2021, mi2020, peral2022, cucurull2023, mukherjee2022, lorentz2023} either through natural language or gestures. One could also include in this category those cases in which the human modifies their behavior to make their intention more easily understandable, e.g., by legible movements~\cite{busch2017}.

Generally speaking, unconscious intention alludes to patterns of human behavior that can be learned (and thus predicted) by the robot, e.g., using neural network-based architectures. Meanwhile, conscious intention is notably less predictable and needs to be elicited at the moment it appears. The distinction between the two types makes it possible to be wary of the response provided by learned inference engines and to opt for direct communication when it is suspected that the human's actions do not follow an unconscious pattern but are the result of a conscious decision.

\subsection{Number of agents: Individual intention VS. Collective intention}\label{sec:individual_VS_collective}

In line with the above, we are going to use the following definitions in robotics:

\begin{itemize}
   \item The \textbf{individual intention} is the intention resulting from a result desired by each agent independently and for the achievement of which they do not take into consideration the collaboration of the other agent.
   \item The \textbf{collective intention} is the common desire that occurs when the human and the robot accept and agree, explicitly or implicitly, to collaborate with each other to achieve a goal or result desired by both parties.
\end{itemize} 

\begin{figure}[t]
    \centering
    \includegraphics[width=0.47\linewidth]{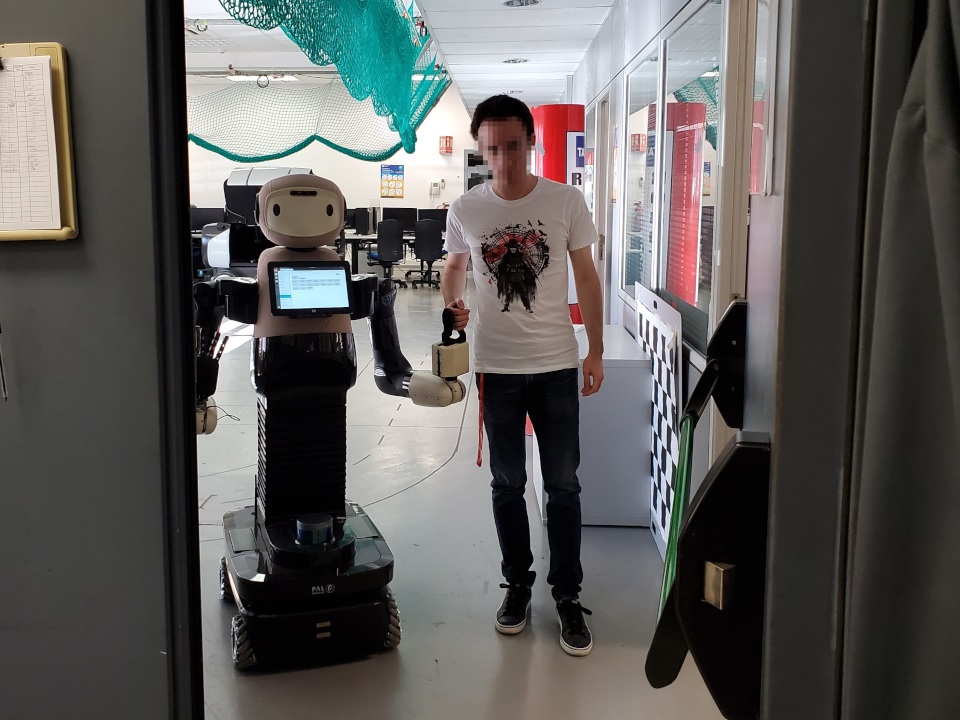}
    \includegraphics[width=0.47\linewidth]{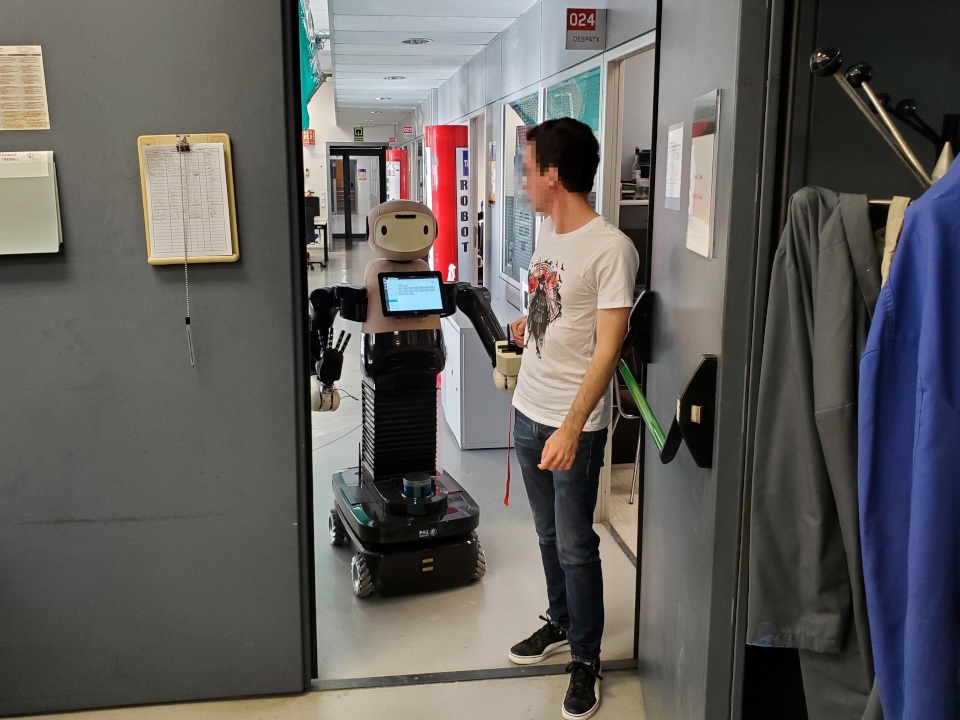}
    %\newline
    \includegraphics[width=0.47\linewidth]{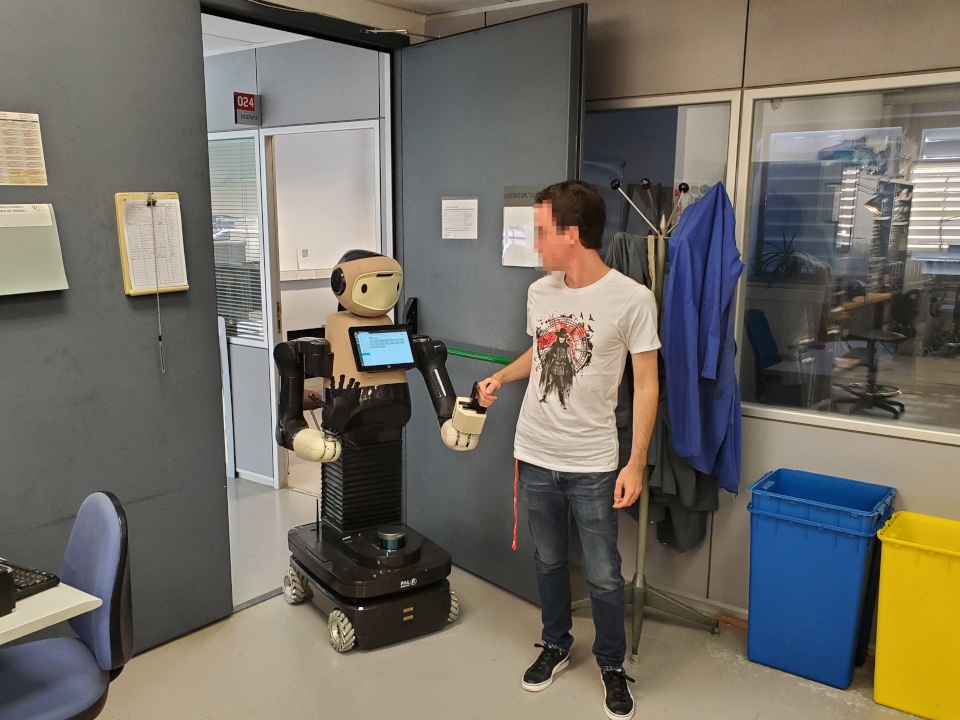}
    \includegraphics[width=0.47\linewidth]{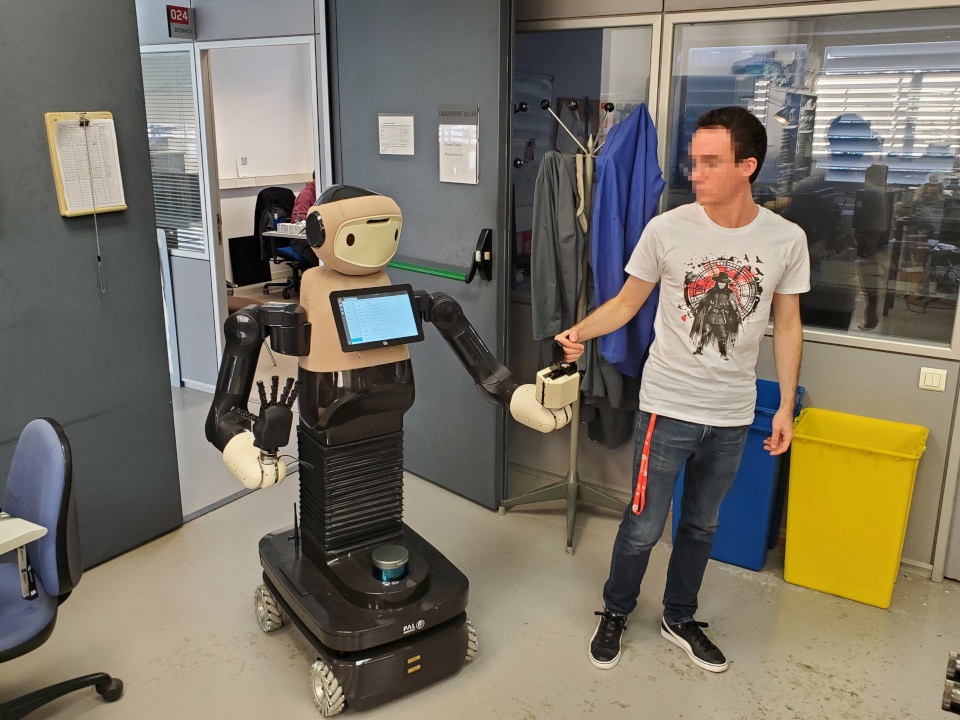}
    \caption{{\bf Human-robot pair navigating through a narrow door.} {\it Top left}: Human-Robot pair navigating side-by-side until they reach a door. {\it Top right}: Human takes the lead but keeps taking into consideration the robot. Robot moves its left arm to allow the human to move forward. {\it Bottom left}: Human waits for the robot to finish passing through the door. Robot rolled up its right arm to reduce its footprint and moved its left arm to make the human to be in front of it. {\it Bottom right}: Human-Robot pair recover their side-by-side configuration.}
    \label{fig:narrow_path}
\end{figure}

Examples of works that consider the human's intention as an individual intention would be those related to autonomous navigation in which the robot must move among humans without being annoying~\cite{ferrer2014b, chadalavada2020}. This is because the human is not being considered to be collaborating with the robot. The same is not true in cases of social navigation~\cite{repiso2020} where human and robot share navigation goals and coordinate their plans so that neither is left behind (see Fig.~\ref{fig:narrow_path}). An interesting case is the one presented in~\cite{chen2019} in which the robot navigates autonomously but taking into account that the humans in its path may have shared intentions between them, i.e., they move in pairs or in groups. This makes it unnecessary to predict the movement of each independent human as they can be considered as a whole. Another example of shared intentions arises in collaborative manipulation tasks~\cite{mavridis2018, alevizos2020, ge2011, li2013} in which an arbitration process~\cite{losey2018} may occur whereby each member of the pair makes the contribution to the task that they deem appropriate.

There are also theoretical works applied to robotics that study the concept of shared intention~\cite{vernon2016, lyons2014} and the need for it to overcome merely instrumental interaction and achieve true collaborations~\cite{vernon2016}, or how the existence of this shared intention together with shared awareness is essential to obtain interactions that are transparent to the human~\cite{lyons2014}.

It is this distinction between individual intention and collective intention that helps explaining phenomena such as mutual adaptation~\cite{nikolaidis2017, nikolaidis2017b}, i.e., the robot adapts to the human's preferences but the human also adapts to the robot's capabilities (typically by moving more slowly or making movements that are more easily interpreted by the robot). This shared intention is useful in generating a compromise between both parties and making it more likely that the human will behave more predictably since this shared intention is known to the robot.

In short, the emergence of a collective intention, typical of those tasks in which there is not only interaction but collaboration between the human and the robot, makes it easier to predict the behavior of the human by knowing the common goal that motivates them to collaborate and knowing that they are willing to help and be helped by the robot because of the commitment underlying this intention. At the same time, the benefits of this type of intention make it advantageous for the robot to encourage its emergence through communication and negotiation with the human.

\subsection{Temporality: Short-term intention VS. long-term intention}

We are going to use the following definitions in robotics:

\begin{itemize}
   \item The \textbf{short-term intention} is the intention which motivates the human's current action or that one which they will perform in the immediate future
   \item The \textbf{long-term intention} is the intention which motivates the human's next actions or tasks, which are dependent on the action or task the human is currently performing or will perform immediately thereafter.
\end{itemize} 

\begin{figure}[t]
    \centering
    \includegraphics[width=0.96\linewidth]{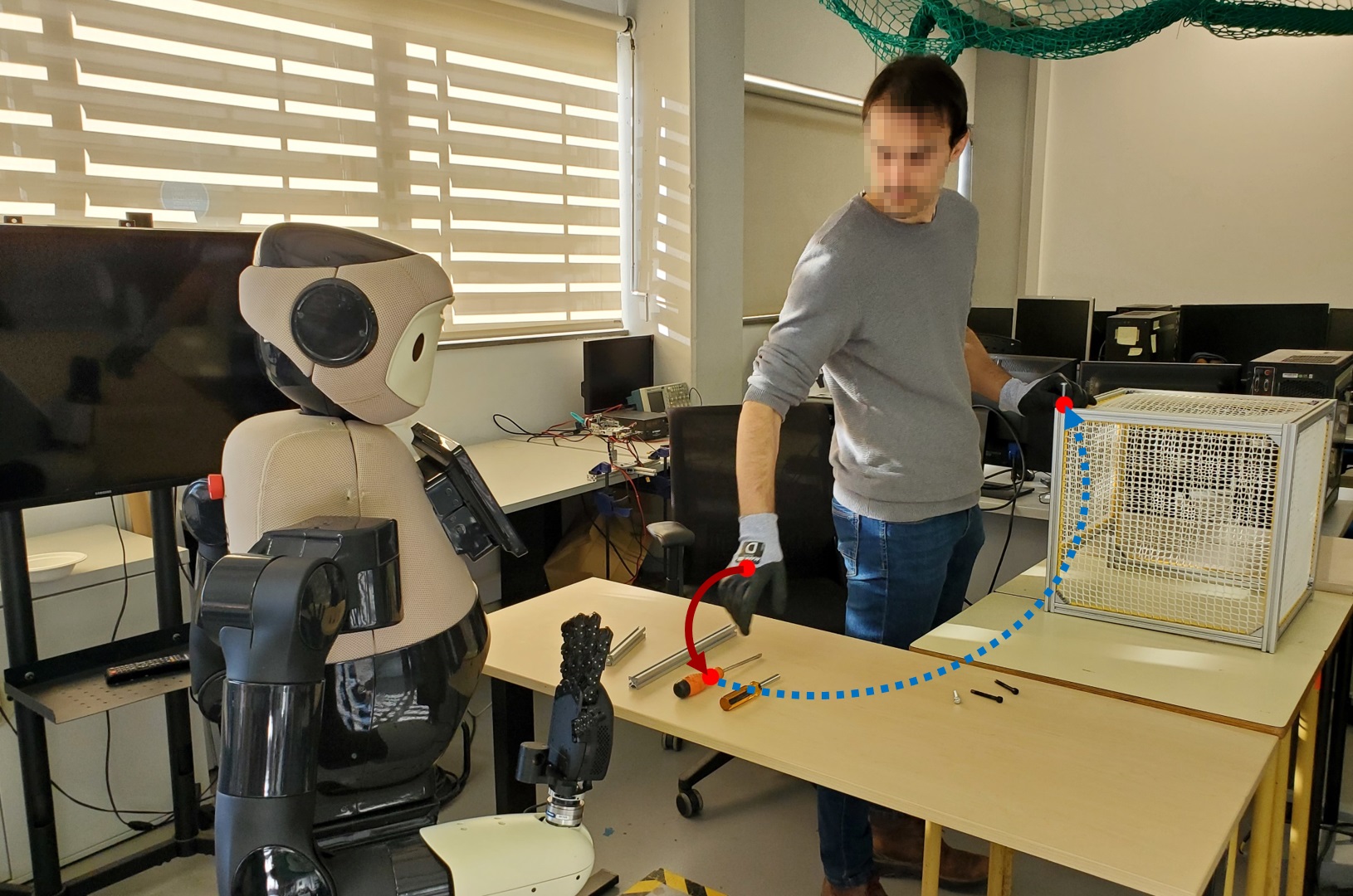}
    \caption{{\bf Collaborative assembly example.} Robot sees that the human is holding a screw and moving his arm towards a screwdriver. It predicts that in the human is going to take the screwdriver in the short term and use it to tighten the screw in the long term. Both predictions allows it to estimate how many time it has to take the next bar and when the shared space will be free.}
    \label{fig:short-VS-long}
\end{figure}

Hence, those works that attempt to predict human intention understood as "the next movements they are going to perform", would be understanding this intention as a short-term intention. This includes multiple movement predictors~\cite{laplaza2022, PIA_HRI, ferrer2014b} with time horizons ranging from less than $1$~$s$ to several seconds in multiple tasks such as hand-over, urban navigation or collaborative transportation. In contrast, those works that seek to understand or predict not only the task currently being performed by the human but also the following tasks or sub-tasks tend to understand the human's intention as a long-term one.

If we focus on assembly tasks, it is possible to find the short-term approach in works like~\cite{unhelkar2018} in which the next human's movement is predicted (with a time horizon of up to $6$~$s$) but not the next tasks to be performed. It is also possible to find the opposite approach in works such as~\cite{zanchettin2018} in which not only the current task being executed by the human is detected but also the next task is predicted and when this will occur so that the robot can be executing other tasks until it is required by the human (see Fig.~\ref{fig:short-VS-long}). Similarly, in~\cite{cheng2021} the human's movement to both complete the current task and the next task is predicted and in~\cite{zhang2022} LfD is used to learn the set and sequence of subtasks associated with each task that the human can perform so that the robot can predict multiple of their next actions. In this way, understanding the human's long-term intention enables the best performance of the human-robot pair by enabling long-term optimization.

At the same time, there is also a relationship between long-term intention and the field of long-term human-robot interaction~\cite{kidd2008, leite2013}, i.e., those interactions that last until after the novelty effect has worn off. This type of interaction occurs when the human has a long-term intention (a desired outcome for which several actions should be performed or maintained for weeks or even months) that they wish to fulfill. The correct understanding of this intention and how it can fluctuate over time is decisive to achieve a successful interaction, allowing the robot to know when to be more insistent or when to just make a reminder.

Generally, increasing the time horizon of the robot's reasoning and trying to understand the causes and long-term consequences of the human's current action is what allows both to improve the performance of the human-robot pair in the short-term and to achieve interactions that are more satisfying for the human in the long-term.

\section{A couple of illustrative examples}\label{sec:use_cases}

Let us choose two different use cases to illustrate how human intention can be analyzed in different ways. First, collaborative search, a task in which there are often occlusions due to the scenario and where distances between the two agents of up to tens of meters can occur. Secondly, collaborative transport of objects, a task performed in close proximity and in which there is a fast exchange of physical forces between the human and the robot.

\begin{figure}[t]
    \centering
    \includegraphics[width=0.96\linewidth]{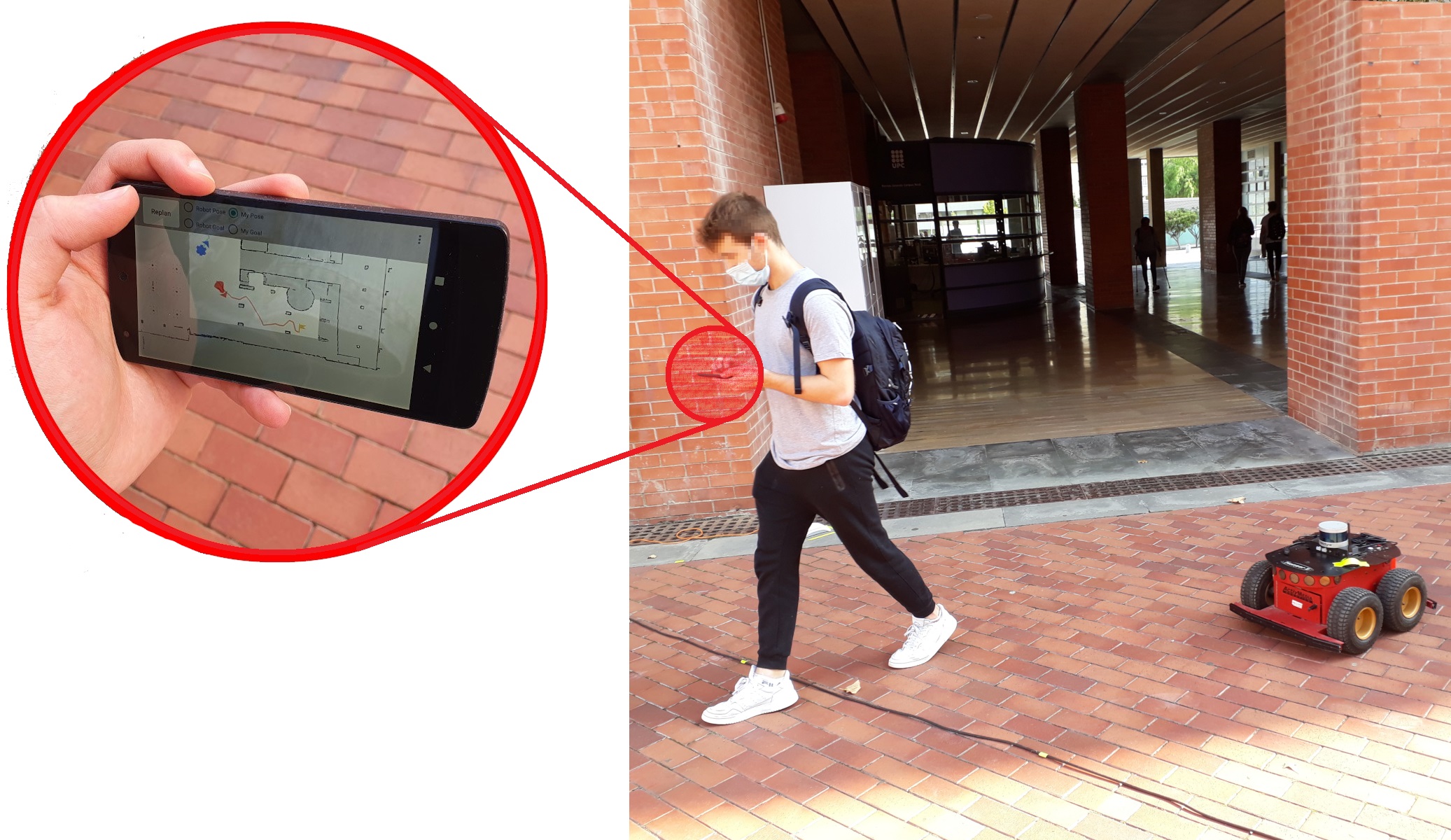}
    \caption{{\bf Human-Robot collaborative search example.} The human is holding an smartphone through which he can communicate explicitly with the robot in a bidirectional manner. The robot can also keep track of the human movement while it can see him.}
    \label{fig:collaborative_search}
\end{figure}

To illustrate the first case, we can use the approach used in~\cite{Dalmasso_ICRA2020} ~\cite{Dalmasso_SORO2023}. In this particular example, they make use of a mobile application through which there are bi-directional communication through both agents despite of the distance~\cite{ROMAN2021} (see Fig.~\ref{fig:collaborative_search}). Thus, the human can indicate in real time the area they want the robot to explore and/or the area they want to explore themselves. With this information the robot calculates a plan for itself and another for the human and communicates both plans to the human through the same application. At the same time, the human can indicate in real time their position so that the robot knows where the human is even when it cannot see them. Likewise, the robot updates in real time its location so that the human knows which areas the robot is exploring.

In this scenario, the human's explicit intention regarding which area they wish to explore is obtained directly through the application whenever the human wishes to express it and their implicit intention is inferred by observing their movements while the human is visible. It is the combination of the two that allows the robot to optimize the search since, in the absence of explicit intention, the robot could plan routes that overlap with areas already explored by the human but unnoticed by the robot. On the other hand, the goal-oriented intention of the human is expressed through the human's indication through the application of the specific place they wish to explore, while the implementation-oriented intention (how meticulously they will explore) can only be obtained by attending to their movement speed. This implementation intention is what allows the robot to reduce to a greater or lesser extent the probability that the searched object is in the areas explored by the human. If the example is analyzed from the perspective of conscious/unconscious intention, the former appears every time the human updates their location or tells the robot a new area to explore, while the latter is not explored in this work but could be understood as the unconscious tendency of the human to accompany the robot or, conversely, to move away from it to cover larger areas depending on their degree of trust in the robot. This would allow the robot to condition its planner to generate routes that optimize the explored area without being too disturbing for the human. At the same time, the existence of a collaborative intention (to find the searched object in the agreed area) is what generates a commitment on the part of the human that allows the robot to update its probability maps believing in the information that the human is giving it.

For the second use case, we can base on~\cite{PIA_HRI, Inferencia_ROMAN2023}. They make a robot and a human to transport an object in a maze with multiple possible paths and several obstacles in the way. The robot is equipped with a force predictor that allows it to estimate the human's desired trajectory in the short term (up to $1$~$s$). In addition, the human can communicate explicitly with the robot through a series of buttons on the handle of the transported object (see Fig.~\ref{fig:collaborative_transportation}).

In this case, the explicit intention is obtained when the human explicitly communicates whether they want to go right or left or if they want to avoid a particular route, while the implicit intention is inferred by the robot from the force exerted by the human. Both types of intent condition the robot's planner but in different ways. The predictor gives the robot a short-term estimate of the route the human would like to follow while the explicit intention directly tells the planner how to plan for the next intersection. On the other hand, the goal-oriented intention of the human refers to the place where they want to take the object while the implementation intention refers to the concrete route that the human wants to follow to take this object. In the cited works, the destination of the transported object is established beforehand. In any case, knowing the goal-oriented intention of the human allows discarding routes because they do not allow reaching the desired destination or simply because they are clearly suboptimal, reducing the planner's computation time. Third, the human's conscious intention is manifested every time they communicate with the robot, while the unconscious intention appears in the tendency or preference of each human to go right or left depending on whether they are right or left-handed or their position with respect to the robot. Taking this into account at the next fork would allow the robot to choose the route that is most comfortable for the human. Finally, taking into account the human's short-term intention to move closer or farther away from the next obstacle without affecting their long-term intention, which would be the complete route to the destination, allows to need only a fine tuning of the robot's movements instead of calling the global planner to recalculate a route that would be remarkably similar to the previous one.

\section{Future challenges}\label{sec:challenges}

All of the use cases presented in the previous sections have either already been realized or may be realizable in the near term given the current state of the field of robotics. However, there remain a number of challenges that will require further research in both the short and long term to be addressed and solved. In this section we address some of these challenges by focusing on extending the understanding of intention not only to humans but also to other machines such as robots or cybernetic avatars.

The two examples presented in the previous section allow to recover the other point of view when analysing the concept of intention previously commented at the beginning of Seccion~\ref{sec:applications} and illustrated in Fig.~\ref{fig:taxonomy-point-of-view}. The commented case of collaborative transport is based on implementing several ways so that the robot can better understand the human's intention. This is the most common approach in the literature. However, the commented case of collaborative search goes one step further. Not only does it seek for the robot to better understand the human's intention by allowing the human to indicate which areas they want the robot to explore and which areas they want to explore by themselves, but it also seeks for the human to better understand the robot's intention by telling the robot which route it is going to follow and which route it expects the human to follow. This second approach, that the robot makes an effort to express its intention, is less common.

Now that we have multiple ways of acquiring human intent in its many facets, one near-term challenge to make these human-robot interactions more natural is to make the robot to also express its intent in one way or another based on what is most useful and relevant in each situation. This not only means to make the robot's movements more legible~\cite{dragan2013, dragan2013b} but to make the robot to really express its intention in an unambiguous way to the human. This would allow a better degree of understanding between both parties by making their behavior more predictable for both, which in turn would lead to better task performance and lower task load and anxiety for the human~\cite{koppenborg2017}.

The next challenge is to make the robot to detect and interpret the intention of another robot. If we pretend to make robots to explicitly communicate among them, a standardized communication code accepted by all should be necessary. This would be challenging taking into account the enormous diversity of capabilities among current robots\footnote{TurtleBot: \url{https://www.turtlebot.com/about/}}\footnote{Pepper: \url{https://us.softbankrobotics.com/pepper}}\footnote{ARI: \url{https://pal-robotics.com/robots/ari/}}\footnote{TIAGo: \url{https://pal-robotics.com/robots/tiago/}}\footnote{Spot: \url{https://support.bostondynamics.com/s/article/About-the-Spot-robot}}~\cite{hri_ivo, mondada2009, saldien2008} which may well be increased in the next years and even decades. The alternative, and probably even more challenging, is to endow the robot with capabilities to recognize the intent of another robot. This requires the first robot to be able to recognize not only the environment and the task being performed by the other robot but also its capabilities. Having done this, it could use the taxonomy outlined in previous sections to detect what is the goal or outcome desired by the other robot as well as how it wishes to implement it. If the interaction with this second robot were regular, the first robot could also try to analyze the other’s routines and detect when they change to know if it can rely on the implicit intention of the other robot indicated by its movements or if it must communicate with it to obtain its intention explicitly.

Focusing on longer-term challenges, it arises the correct understanding of how the comprehension of each agent’s intention affects in the whole decision-making process, including the negotiation that inevitably arise when their respective intentions do not match or the assignment of roles when one of the agent's intention is dominant over the others. While this negotiation process has been slightly studied in robotics~\cite{thomas2018, chandan2019, moon2021, aydougan2021}, how the intensity of each agent’s intention turns out to be determinant in knowing how or how far to negotiate remains completely unexplored. The same happens with the assignment of roles, for example, in who is going to be the leader or the follower in a cooperative task; where the correct understanding of each partner's intention can be fundamental.

The correct understanding and use of several of the classifications listed in the previous sections can be really useful in enabling the robot to detect cases such as when the human is making a joke or resorting to irony. Or even worse, when they are lying, situation almost unexplored in robotics since we always consider the goodness of human as if they were always willing to collaborate with the robot. Discrepancies between conscious and unconscious intent or between implicit and explicit intent could guide the robot to learn to recognize these cases and even know how to deal with them. Likewise, these aspects could enable the robot to face the challenge of being able to display irony-like behaviors or to know when it is appropriate to make a joke, taking it a step closer to being the robot accepted as a human companion.

Taking into account the occurrence of collective intentions can help the robot by more easily modeling the intention of a group of people as a whole rather than as several independent individual intentions. However, the problem becomes more complicated when this group is composed of both humans and robots. Should the robot interpret the intent of other robots collaborating with humans as if they were other humans? Should these robots indicate their intent in a way that is easily interpretable to both the humans they are working with and other robots that may perceive them? Or, should them use different communication channels since they are going to have potentially different and complementary capabilities from humans? Many questions can be considered being the answer to each of them a challenge in itself.

Finally, the emergence of cybernetic avatars~\cite{ishiguro2021} (robots that are fully or partially teleoperated by a human operator when desired) and their possible proliferation in the years to come adds an extra level of difficulty. These partially teleoperated robots make possible the appearance of mixed intentions as a combination of the intention of the human and that of the avatar. The latter follows its own intention when it is operating autonomously. However, when teleoperated by the human, the avatar must interpret their intentions, with this interpretation being more or less transparent depending on the degree of control and expressiveness that the human has through the avatar. What are the implications of these considerations or how this intention should be shown in such a way that it is easy for the interlocutor to differentiate the contribution of each agent are questions that are currently unanswered. Add to this the option that multiple humans could operate the same avatar, and the potential challenges grow exponentially.

\section{General discussion}\label{sec:discussion}

Human intention has been classified into various categories, indicating in each case the two extreme types that this intention can have. However, it is necessary to remember that typically they are not two isolated possibilities but of a continuum that goes from one extreme to the other. Except in the case of the goal-oriented/implementation distinction, in the other categories the intention can take intermediate values between the two extremes. This is particularly noticeable in the case of the temporality classification, where it is totally context-dependent what we consider to be short-term and what means long-term. Likewise, the different classifications are not mutually exclusive, since the same intention can be observed from different prisms and, therefore, belong to several categories. Proof of this is that multiple works have been cited as examples in several categories at the same time.

In this regard, it is worth emphasizing that Fig.~\ref{fig:categories-relation} attempts to show some of these relationships, although it is only an illustrative representation. Therefore, it should not be interpreted that any of the categories analyzed here result from a linear combination of others, but only that there are dependencies between them. In turn, these dependencies do not necessarily occur in all cases, but are the tendency that, by their nature, they tend to present.

Regarding the distinction between implicit/explicit and unconscious/conscious intention, in the psychological literature both concepts can be found intermingled~\cite{ajzen2015, wegner1998, sheeran2013}, defining as implicit that intention of which the human is not fully aware. At the same time, in robotics we do find specific cases in which explicit intent is defined as that which is directly stated and implicit as the one that must be inferred~\cite{chadalavada2015, che2020, PIA_HRI}. Because this work does not seek to focus on psychological studies but to use them as a basis for explaining different use cases in robotics, we have chosen to distinguish between the two categories to be able to accommodate a larger number of cases.

Finally, it is worth mentioning that it is possible to consider other ways of classifying human intention than the one presented in this paper. One of them could be to consider a hierarchical relationship with a main intention and that this is composed of multiple sub-intentions and these in turn composed of more atomic intentions in a way similar to how human goals are classified in~\cite{chulef2001}. However, we consider that the goal-oriented/implementation and short-term/long-term intention classifications could explain such taxonomies. It should also be pointed out that, to the best of our knowledge, no work has been found in the field of psychology that attempts to present a complete taxonomy of human intention, making the present article potentially of interest not only in robotics but also for future psychological studies. 

In any case, the aim of this article is not so much to be scrupulous and show all the possible types of intention but to ask certain questions that in fields such as psychology have managed to answer but that in robotics we may be taking for granted, such as: what is intention? In turn, we hope that this work will serve as a stepping stone for other technical researchers who are taking their first steps in the field of psychology and that it will serve as a small summary of what can be found in the literature and as a link to multiple articles with which to delve deeper into each related topic.

\section{Conclusions}\label{sec:conclusions}

In this article we have drawn on decades of psychological research to provide a universally accepted definition of the concept of human intention. Also based on psychological and interpersonal communication studies, we have defined five different ways of classifying human intention and analyzed each of them separately as well as the relationships that appear among them. This has allowed us to see how multiple works in the field of robotics can fit several classifications at the same time. Finally, by means of two specific use cases, we have also verified how taking into account the multifaceted nature of human intention can help to gain a better understanding of human behavior. We believe that this work can inspire other researchers to both extend our classification and to approach their work from other perspectives that will allow them to achieve higher levels of performance and more human-like human-robot interactions.

\backmatter

\section*{Declarations}\label{sec:declarations}

%Some journals require declarations to be submitted in a standardised format. Please check the Instructions for Authors of the journal to which you are submitting to see if you need to complete this section. If yes, your manuscript must contain the following sections under the heading `Declarations':

\bmhead{Funding}

Work supported under the European project CANOPIES (H2020-ICT-2020-2-101016906) and by JST Moonshot R \& D Grant Number: JPMJMS2011-85. The first author acknowledges Spanish FPU grant with ref. FPU19/06582.

\bmhead{Competing interests}

There are none potential conflicts of interest that could bias the evaluation or results of our research.

\bmhead{Data availability}

We do not analyse or generate any datasets, because our work proceeds within a theoretical approach.

% \bmhead{Ethics approval}

% All the experiments reported in this document have been performed under the approval of the ethics committee of the Universitat Politècnica de Catalunya (UPC) in accordance with all the relevant guidelines and regulations (ID: 2021.10).

% \bmhead{Consent to participate}

% All the volunteers who participated in this study have signed an informed consent form accepting to participate in the study.

% \bmhead{Consent for publication}

% All the volunteers who participated in this study have signed an informed consent form accepting to publish the anonymously obtained data.

%\item Availability of data and materials
%\item Authors' contributions

\bigskip

\bibliography{./sn-bibliography.bib}% common bib file

\end{document}